\documentclass{article}

    \PassOptionsToPackage{numbers, compress}{natbib}

    \usepackage[preprint]{neurips_2024}

\usepackage[utf8]{inputenc} 
\usepackage[T1]{fontenc}    
\usepackage{hyperref}       
\usepackage{url}            
\usepackage{booktabs}       
\usepackage{amsfonts}       
\usepackage{nicefrac}       
\usepackage{microtype}      
\usepackage{xcolor}         
\usepackage{amsmath}
\usepackage{graphicx} 
\usepackage{subfig}

\title{Understanding the Limitations of Diffusion Concept Algebra Through Food}

\author{E. Zhixuan Zeng \qquad Yuhao Chen \qquad Alexander Wong \\
University of Waterloo\\
{\tt\small \{ezzeng, yuhao.chen1, alexander.wong\}@uwaterloo.ca}
}

\begin{document}
\maketitle
\begin{abstract}

Image generation techniques, particularly latent diffusion models, have exploded in popularity in recent years. Many techniques have been developed to manipulate and clarify the semantic concepts these large-scale models learn, offering crucial insights into biases and concept relationships. However, these techniques are often only validated in conventional realms of human or animal faces and artistic style transitions. The food domain offers unique challenges through complex compositions and regional biases, which can shed light on the limitations and opportunities within existing methods. Through the lens of food imagery, we analyze both qualitative and quantitative patterns within a concept traversal technique. We reveal measurable insights into the model's ability to capture and represent the nuances of culinary diversity, while also identifying areas where the model's biases and limitations emerge.

\end{abstract}    
\section{Introduction}
\label{sec:intro}


In recent years, latent diffusion models have gained significant traction, heralding a new era of remarkably high-quality generated images. With this, considerable effort has been made toward identifying better ways to examine, disentangle, and traverse through higher-level concepts learned during training. Recent works include those focused on image concept manipulation and control
\cite{galTextualInversion2022,gandikotaConceptSlidersLoRA2023,avrahamiBreakASceneExtractingMultiple2023,liuCompositionalVisualGeneration2023}, 
model unlearning 
\cite{kumariAblatingConceptsTexttoImage2023,gandikotaErasingConceptsDiffusion2023}, 
concept disentanglement 
\cite{jinClosedLoopUnsupervisedRepresentation2024,liGetWhatYou2024,motamedLegoLearningDisentangle2023,rassinLinguisticBindingDiffusion2023}, 
along with ways to better understand the underlying concept representation, such as through concept decomposition \cite{cheferHiddenLanguageDiffusion2023}, or identifying semantic directions 
\cite{parkUnsupervisedDiscoverySemantic2023,liSelfDiscoveringInterpretableDiffusion2023,haasDiscoveringInterpretableDirections2023,brackSEGAInstructingTexttoImage2023,dalvaNoiseCLRContrastiveLearning2023,kwonDiffusionModelsAlready2022,wangConceptAlgebraScoreBased2023}. 
These advancements not only enhance control over the model's outputs but also reveal hidden biases and relationships within the concepts that are learned. However, their evaluation is often limited to a narrow set of commonly studied areas like human and animal faces \cite{parkUnsupervisedDiscoverySemantic2023,liSelfDiscoveringInterpretableDiffusion2023,haasDiscoveringInterpretableDirections2023,brackSEGAInstructingTexttoImage2023,dalvaNoiseCLRContrastiveLearning2023,kwonDiffusionModelsAlready2022,wangConceptAlgebraScoreBased2023} or artistic style transfer \cite{wangConceptAlgebraScoreBased2023}. This narrow focus limits our understanding of their behaviour when applied to more complex or less structured data.

Enter the domain of food imagery — a vastly under-explored area in the context of latent diffusion models. Food presents an intriguing case study for several reasons. First, it often involves compositions of multiple subjects, such as several dishes on a table or plates arranged with a main dish, side dishes, and even decorative arrangements. Second, some foods can exhibit strong regional influences, with distinct appearances and dishes in different cuisines. These influences can provide a convenient avenue for understanding how the model deals with biases and entanglement between food dishes and regions.

The diversity in food types — from regional specialties to various preparation methods — also introduces a high degree of variability in visual characteristics. These characteristics, along with the universal familiarity with food, makes it an ideal candidate for exploring the limitations of semantic directions and concept decomposition techniques.

In this paper, we explore one recent method, called Concept Algebra \cite{wangConceptAlgebraScoreBased2023}, within the context of food image generation. We will focus on identifying limitations, particularly in the context of causal separability \cite{wangConceptAlgebraScoreBased2023} and feature representation, and propose metrics useful for model understanding and prompt engineering.
\section{Background}
\label{sec:background}

Semantic directions refers to vectors on some learned latent space corresponding to a semantic concept, where adding or subtracting that vector would result in meaningful semantic change. This idea has existed for many years, with the famous example being ``King - male + female = Queen" \cite{mikolovLinguisticRegularitiesContinuous2013}.

Concept Algebra \cite{wangConceptAlgebraScoreBased2023} is a recent technique applying semantic directions to diffusion models for controlled image generation. To make it easier to find these semantic directions, concepts are separated into different ``sets", each with a projection into a subspace. Traversal through this subspace maximizes changes relevant to the target ``set" of concepts, while minimizing changes in other concepts. 

For example, given two concept sets $\mathcal{Z}$ and $\mathcal{W}$ referring to ``style" and ``content" respectively, an image may be composed of a ``style" concept $\text{Z}$ and a ``content" concept $\text{W}$ (eg. ``a mall in photorealistic style"). Identifying a direction in the ``style" subspace corresponding to a change from ``photorealism" to ``painting" would allow the user to generate a ``painting" styled image, while maintaining the same content as the original prompt (eg. ``a mall in painted style").

To identify the subspace for $\mathcal{Z}$, a set of prompts varying in $\mathcal{Z}$ but constant in $\mathcal{W}$ is used. The constant content concept ($\text{W}_0$) should be easy to generate for all $\mathcal{Z}$. The model evaluates the latent vector conditioned on each prompt.
The top $K$ eigenvectors are identified and used as the unit vectors for the ``style" subspace $\mathcal{R}_Z$.  $\text{proj}_z$ defines projection from the latent space into the ``style" subspace. The image edit can then be performed through:

\begin{equation}
s_{edit} \leftarrow (\mathbb{I} - \text{proj}_z)s[x_\text{orig}] + \text{proj}_z s[x_\text{new}]
\end{equation}

\noindent where $s[x]$ is the latent vector at each time step conditioned on a prompt $x$, $x_\text{orig}$ is a prompt containing target ``content" concept ($\text{W}_t$) and ``style" concept $\text{Z}_0$ that pairs easily with $\text{W}_t$, while $x_\text{new}$ contains some target ``style" concept ($\text{Z}_t$) and the same $\text{W}_0$ used in calculating the ``style" subspace. $\mathbb{I}$ is the identity matrix. The edit ideally subtracts the ``style" in $x_\text{orig}$ and adds in the ``style" in $x_\text{new}$, without changing the ``content" from $x_\text{orig}$.

\section{Experimental Setup}
\label{sec:experimental_setup}

Our goal is to explore Concept Algebra's limitations in the context of food imagery and discover if the concept subspace offers any benefits beyond image editing. Our experiments primarily focus on two concept sets: ``cuisine region" as ``style" concepts and ``food ingredients" as ``content" concepts. Other ``style" sets, such as ``fast-food" vs ``fine-dining" and food preparation techniques (eg. fried, boiled) is also explored visually.

\begin{figure}
    \centering
    \includegraphics[width=\linewidth]{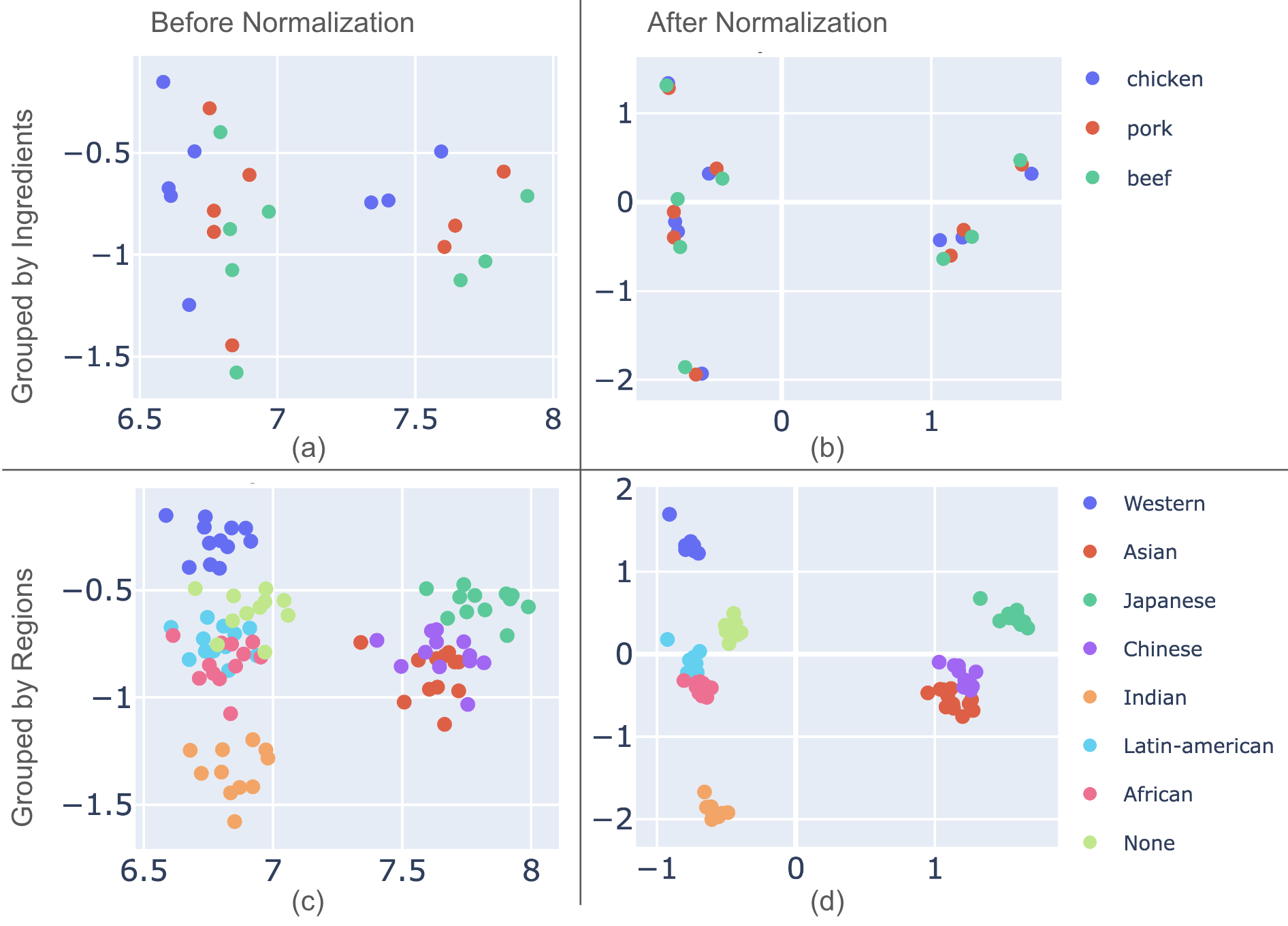}
    \caption{Prompts containing different food ingredients and cuisine regions plotted in a ``cuisine regions" subspace. (a) and (c) shows the vectors colored by ingredients and region respectively. (b) and (d) shows the same vectors after normalizing by the mean and variance of each ingredient distribution.}
    \label{fig:before_after_norm}
\end{figure}

\textbf{Subspace Visualization and Clustering:} 
First, we visualize the projection of many different prompt results into the ``style" subspace (Figure \ref{fig:before_after_norm}c). These prompts vary in both ``content" and ``style". 

We define a cluster of subspace vectors as a collection of outputs conditioned on the same ``style". Because the subspace should maximize changes only in ``style", we hypothesize that intra-cluster distances should ideally be closer together than inter-cluster distances. We also hypothesize that inter-cluster distances may provide meaningful information on the perceived similarity between each ``style" (ie. Chinese cuisine may be closer to Japanese Cuisine than American cuisine). Therefore, after collecting the subspace vectors, we calculate the inter-cluster distances using the Jensen-Shannon distance \cite{lin1991divergence}, a distribution distance measure based on KL divergence.
 We further evaluate clustering quality with the silhouette score \cite{rousseeuw1987silhouettes}, a measure comparing intra-cluster distance with inter-cluster distance. The score varies in range [-1, 1], where a higher score means tighter, more distinct clusters.


\textbf{Causal Separability:} The original work \cite{wangConceptAlgebraScoreBased2023} states causal separability as a prerequisite for the technique, meaning the way the two sets of concepts effect the final output must not depend fundamentally on some interaction between the two concepts. However, this definition is not truly clear-cut. Concept entanglement is a well-observed phenomenon in diffusion models \cite{jinClosedLoopUnsupervisedRepresentation2024,liGetWhatYou2024,motamedLegoLearningDisentangle2023}, and learned biases may cause concepts which are theoretically causally separable difficult to separate in practice.


In our experiments, we observed that a change in ingredient may result in a similar change in the region subspace vector across all regions (Fig. \ref{fig:before_after_norm}a). We hypothesize that a larger, more consistent change in the ``style" subspace given ``content" concepts suggests limited causal separability, while a weaker correlation to changes in ``content" concepts suggests stronger causal separation. This can be visualized by normalizing the subspace vectors by the mean and variance of each ingredient distribution (Fig. \ref{fig:before_after_norm}b and Fig. \ref{fig:before_after_norm}d). The change can also be measured by the change in silhouette score before and after normalization.

\section{Observations}


\begin{figure}
      \centering
      \includegraphics[width=0.5\linewidth]{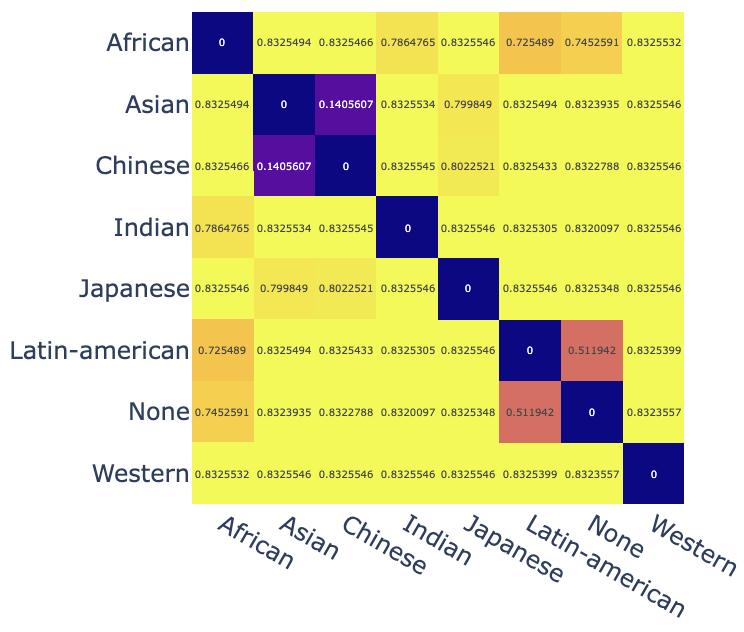}
      \caption{Jensen-Shannon distance, a distribution distance metric, between different cuisine region clusters}
      \label{fig:nonorm_cuisine_chicken_js}
\end{figure}

\subsection{Measurable Observations}
\label{sec:measurable_observations}

\textbf{Subspace Visualization and Clustering:}
An example visualization of the projected subspace vectors can be seen in Fig. \ref{fig:before_after_norm}c, with the Jensen-Shannon distances between each cluster visualized in Fig. \ref{fig:nonorm_cuisine_chicken_js}. 
, where we see that the cluster containing ``Chinese cuisine" is very close to the cluster for ``Asian cuisine". This shows that, using these subspace vectors, we may be able to easily identify some biases and relationships between categories in the concept set. 

\begin{figure}
    \centering
    \includegraphics[width=\linewidth]{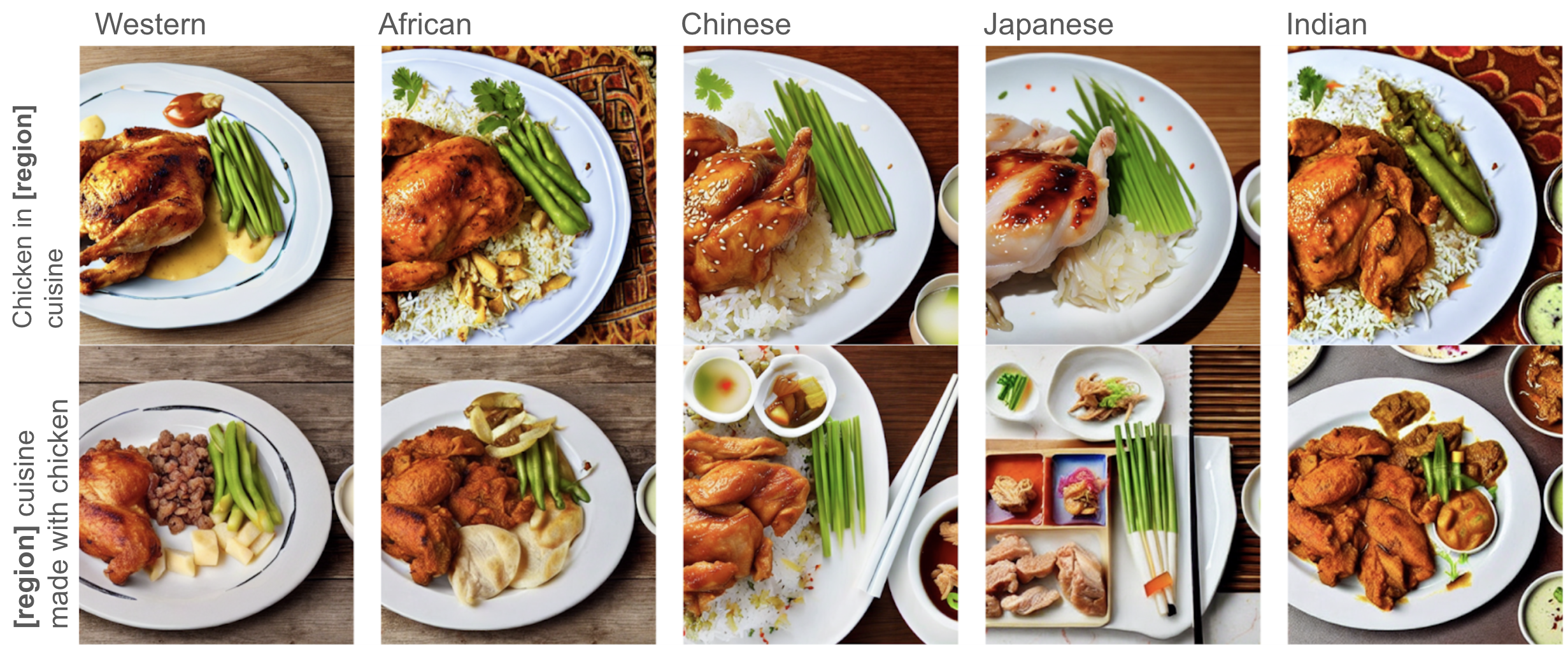}
    \caption{Comparing the result of different prompt wording. Visually, the first row has limited variation in composition and the cut and shape of the chicken compared to the second row. Numerically, silhouette scores in the first row are 0.33 and 0.61 for non-normalized and normalized subspace vectors respectively, while the second row is 0.61 and 0.63}
    \label{fig:region_prompt_compare}
\end{figure}

\textbf{Causal Separability:} Through testing, we found that the difference in Silhoutte Score before and after normalization can change significantly based on prompt wording. Figure \ref{fig:region_prompt_compare} shows two sets of experiments: one generated using prompts formatted as ``{ingredient} in {Region} cuisine" (eg. ``food in Chinese cuisine"), and the second using ``food in {region} cuisine made with {ingredient}"

With the same ``content" and ``style" concepts, the silhouette scores for the first prompt format is 0.33 before normalization and 0.61 afterwards, while the second prompt is 0.61 and 0.63 for non-normalized and normalized subspace vectors respectively. This much smaller difference shows a significant decrease in the influence ``ingredients" have in this new ``cuisine region" subspace.

To further explore this phenomenon, instead of generic ingredients common to all cuisines, we select foods with strong regional biases as ``content" concepts, including stirfry, sushi, bread, pasta, dimsum, curry, and tacos. Given the strong regional biases, we expect that changes in ``content" would heavily influence the ``cuisine region" subspace. This is verified, as when we attempted to transform ``bread" into ``Japanese cuisine", we ended up with silhouette scores of -0.17 and 0.54 for non-normalized and normalized subspace vectors respectively.

We also observed examples where the silhouette scores were low both before and after normalization. When we set the ``style" concepts to be different cooking techniques (eg. deep-fried, steamed, baked, raw), we
saw very little variation in the image outputs between the different categories. Correspondingly, the clusters in the ``cooking techniques" subspace were not very distinct, with ``Raw chicken" receiving a silhouette score of 0.11 both before and after normalization. This suggests that the poor image results may not be due to separability issues, but perhaps the space itself is not very expressive.

To summarize: Large difference between normalized and non-normalized silhouette scores in the ``style" subspace suggests issues with causal separability. Tracking silhouette scores can aid in finding better prompt wording. Low silhouette scores results in lower variation in generation outputs and poor image results.

\subsection{Visual Observations}

\begin{figure}
    \centering
    \subfloat[\label{fig:poached_chicken}``Poached Chicken"]{
      \includegraphics[width=0.3\linewidth]{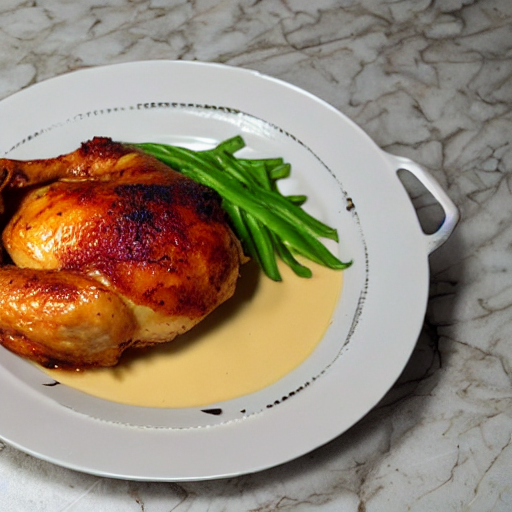}
     }
    \subfloat[\label{fig:boiled_chicken}``Boiled Chicken"]{
      \includegraphics[width=0.3\linewidth]{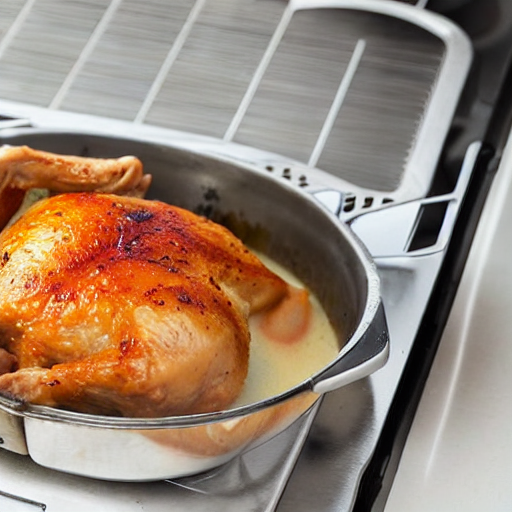}
    }
    \subfloat[\label{fig:chinese_chicken}``food in Chinese cuisine made using chicken"]{
      \centering
      \includegraphics[width=0.3\linewidth]{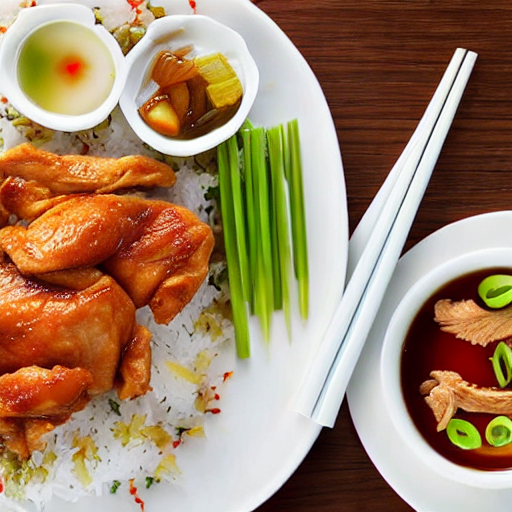}
    }
    
    \caption{Examples of style changes where peripheral elements rather than the main element were changed significantly}
    \label{fig:chicken_img}
    
\end{figure}

Our visual analysis within the food imagery domain has also yielded insightful observations. In the ``cooking techniques" example (Fig. \ref{fig:poached_chicken}, \ref{fig:boiled_chicken}), we observed that oftentimes, the food itself was not transformed to fit its ``boiled" or ``poached" characteristics. Instead, peripheral elements within the image changed. This is not an isolated occurrence. For instance, in for Asian cuisine, chopsticks are often added, along with small dishes of other common ``asian food" such as rice (Fig \ref{fig:chinese_chicken}).

This observation is difficult to uncover in other common domains of exploration, as those domains either usually contain only a single subject (eg. people), or requires changing the entire image to take effect (artistic styles). In contrast, food images often contains multiple subjects, but does not require changing the entire art style to move from one cuisine region to another.

\begin{figure}
    \centering
    \subfloat[\label{fig:chicken_burger}``chicken in fast food"]{
      \includegraphics[width=0.3\linewidth]{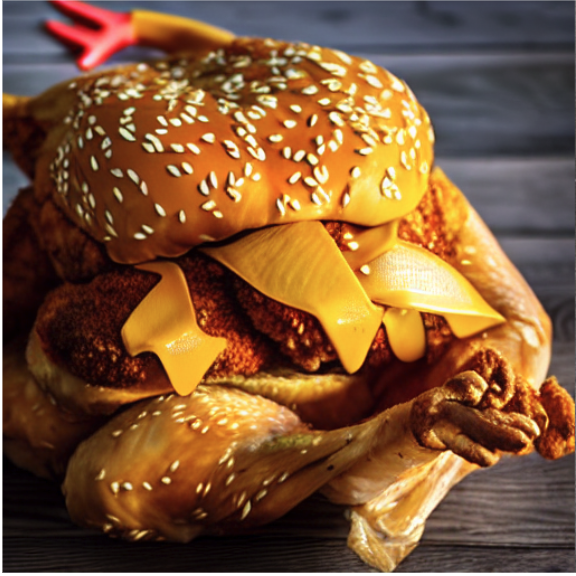}
     }
    \subfloat[\label{fig:sushi_bread}``bread in Japanese cuisine"]{
      \includegraphics[width=0.3\linewidth]{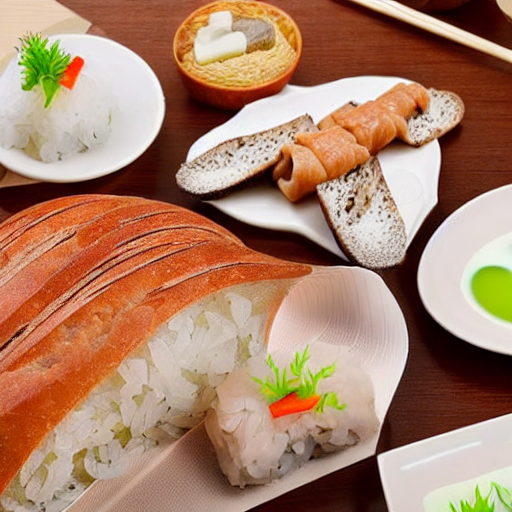}
    }
    \subfloat[\label{fig:sushi_sandwich}``sandwich in Japanese cuisine"]{
      \includegraphics[width=0.3\linewidth]{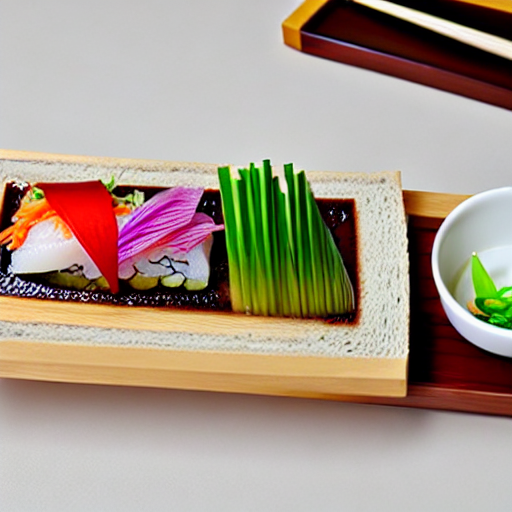}
    }
    \caption{Examples where the texture of the original ingredient was placed in the form of some ``default" food item in the target ``style" concept}
    \label{fig:texture_reliance_examples}
    
\end{figure}

Furthermore, our exploration revealed a tendency of diffusion models to prioritize alterations in the visual features of an image over the accuracy of the subject's transformation into a proper food item. A striking example of this can be observed when Concept Algebra is used to transform ``bread in western cuisine" into Japanese cuisine. The result is an image where sushi appears with a top texture reminiscent of bread (fig. \ref{fig:sushi_bread}). Similarly, a sandwich transformed into Japanese cuisine sees the bread being repurposed as a plate for sushi (fig. \ref{fig:sushi_sandwich}). Chicken transformed into fast-food has also yielded a chicken-shaped burger (fig. \ref{fig:chicken_burger}).

Two problems may be contributing to this phenomenon. First, this demonstrates a higher focus on textural features as opposed to semantic meaning. This is potentially due to using post-unet features as the latent space to construct the ``style" subspace. Future work may look into experimenting with vectors at the middle of the U-net, as done in \cite{haasDiscoveringInterpretableDirections2023,liSelfDiscoveringInterpretableDiffusion2023,parkUnsupervisedDiscoverySemantic2023}, 
or CLIP space \cite{cheferHiddenLanguageDiffusion2023,oldfieldPartsSpeechGrounded2023}.

Second, the prompts used in constructing the ``style" subspace use a single prompt per region. As such, those regions may become overly represented by the result of that single prompt. Existing biases (eg. Japanese cuisine = sushi) may be amplified the use of the single exemplar. 
It is also possible that the model has learned only a very limited representation of the concept and cannot express greater variation.

\section{Conclusion}

Through applying concept algebra to food imagery, we've revealed patterns and metrics unique to its complexities. We quantified distances between cuisine clusters, identifying important biases and relationships between categories. Through measuring cluster distinctiveness, we can estimated separability between two concept sets of interest. By identifying improvements in those metrics, we can aid in prompt engineering through quantitatively measuring prompt performance.
 
Visually, we found a model bias towards textural over semantic accuracy, leading to transformations that reveal strong biases in the model, potentially emphasized by the concept algebra technique. Instances where peripheral elements in images changed, instead of the food items themselves, reveal another area of improvement.

Concept Algebra, and many other similar techniques exploring a model's latent representation, are vital to developing a more nuanced, accurate, and culturally comprehensive understanding of the large-scale generative models in use today. However, the limitations of these techniques can only be better understood through exploring diverse testing domains. The nuanced exploration of the food imagery domain promises to enrich future research, advancing the depth and accuracy of generative models in capturing complex subjects.
{
    \small
    \bibliographystyle{unsrtnat}
    \bibliography{main}
}


\end{document}